\documentclass[conference]{IEEEtran}
\IEEEoverridecommandlockouts
\usepackage{cite}
\usepackage{amsmath,amssymb,amsfonts}
\usepackage{algorithmic}
\usepackage{graphicx}
\usepackage{textcomp}
\usepackage{xcolor}
\usepackage{url}
\usepackage{bbm}
\usepackage{mathrsfs}
\def\BibTeX{{\rm B\kern-.05em{\sc i\kern-.025em b}\kern-.08em
    T\kern-.1667em\lower.7ex\hbox{E}\kern-.125emX}}
\begin{document}

\title{Deep Fusion: An Attention Guided Factorized Bilinear Pooling for Audio-video  Emotion Recognition\\
}

\author{\IEEEauthorblockN{Yuanyuan Zhang, Zi-Rui Wang, Jun Du}
\IEEEauthorblockA{
National Engineering Laboratory for Speech and Language Information Processing,\\
University of Science and Technology of China, Hefei, China\\
Email: zyuan@mail.ustc.edu.cn, cs211@mail.ustc.edu.cn, jundu@ustc.edu.cn}
}
\maketitle

\begin{abstract}
Automatic emotion recognition (AER) is a challenging task due to the abstract concept and multiple expressions of emotion. Although there is no consensus on a definition, human emotional states usually can be apperceived by auditory and visual systems. Inspired by this cognitive process in human beings, it's natural to simultaneously utilize audio and visual information in AER. However, most traditional fusion approaches only build a linear paradigm, such as feature concatenation and multi-system fusion, which hardly captures complex association between audio and video. In this paper, we introduce factorized bilinear pooling (FBP) to deeply integrate the features of audio and video. Specifically, the features are selected through the embedded attention mechanism from respective modalities to obtain the emotion-related regions. The whole pipeline can be completed in a neural network. Validated on the AFEW database of the audio-video sub-challenge in EmotiW2018, the proposed approach achieves an accuracy of 62.48\%, outperforming the state-of-the-art result. 
\end{abstract}

\section{Introduction}
Emotions play an important role in human communications~\cite{cowie2001emotion} and successfully detecting the emotional states has practical importance in sociable robotics, medical treatment, education  quality evaluation and many other human-computer interaction systems. Audio and video, more specifically, the speech and facial expressions are two kinds of most powerful, natural and universal signals for human  beings to convey their emotional states and intentions~\cite{darwin1998expression},~\cite{tian2001recognizing}. According to the linguistics and physiology, emotion changes voice characteristics and linguistic contents in speech and moves the  facial muscles~\cite{facialexp}. 

Considering its importance, several challenges were held to facilitate the research on emotion recognition. As one of the most popular benchmarks for this task, the emotion recognition in the wild (EmotiW) challenge~\cite{dhall2018emotiw} has been held successfully for 5 years.  The audio-video sub-challenge aims to identify people's emotions based on audio-video clips in the AFEW database~\cite{dhall2012collecting}, which is collected from films and TV series to simulate the real world. The wide illumination range and occlusions make the task challenging.

For facial emotion recognition (FER), the traditional approaches usually consist of three steps: First, a face image is detected from an input image, and the facial components (e.g., eyes and nose) or landmarks are identified from the face region. Next, various spatial and temporal features are extracted from these facial components. Finally, based on the extracted features. a classifier such as support vector machine (SVM), random forest, is trained to produce recognition results~\cite{ko2018brief}. Different from using handcrafted features in the traditional approaches, deep learning based FER systems 
adopt convolutional neural network (CNN) to directly learn the task-dependent features from raw face images. The feature maps are combined by following fully-connected layers or global pooling layer for the classification. In~\cite{hu2017learning}, the authors presented the supervised scoring ensemble (SSE) with deep CNNs to provide sufficient supervision information to the shallow layers in CNN. In~\cite{fan2018video}, deeply-supervised CNN (DSN) architecture with a series of side-output layers was proposed to combine the shallow and the deep layers together to achieve a complementary effect. For the face image sequences in videos, the  long short-term memory (LSTM) and 3D convolutional neural network (3D CNN) are widely used to analyze the temporal features in the previous audio-video sub-challenges in EmotiW~\cite{tran2015learning},~\cite{vielzeuf2017temporal},~\cite{kim2017multi}.  Intuitively, not all the frames in a video contain emotion information due to the sparse expression of emotion. In order to adapt the weight of each frame to the final classifying,  according to its importance to emotion, we propose an attention mechanism to detect the emotion-dependent frames in the face image sequences.

\begin{figure*}[!ht]
\centering
\includegraphics[width=1.0\linewidth]{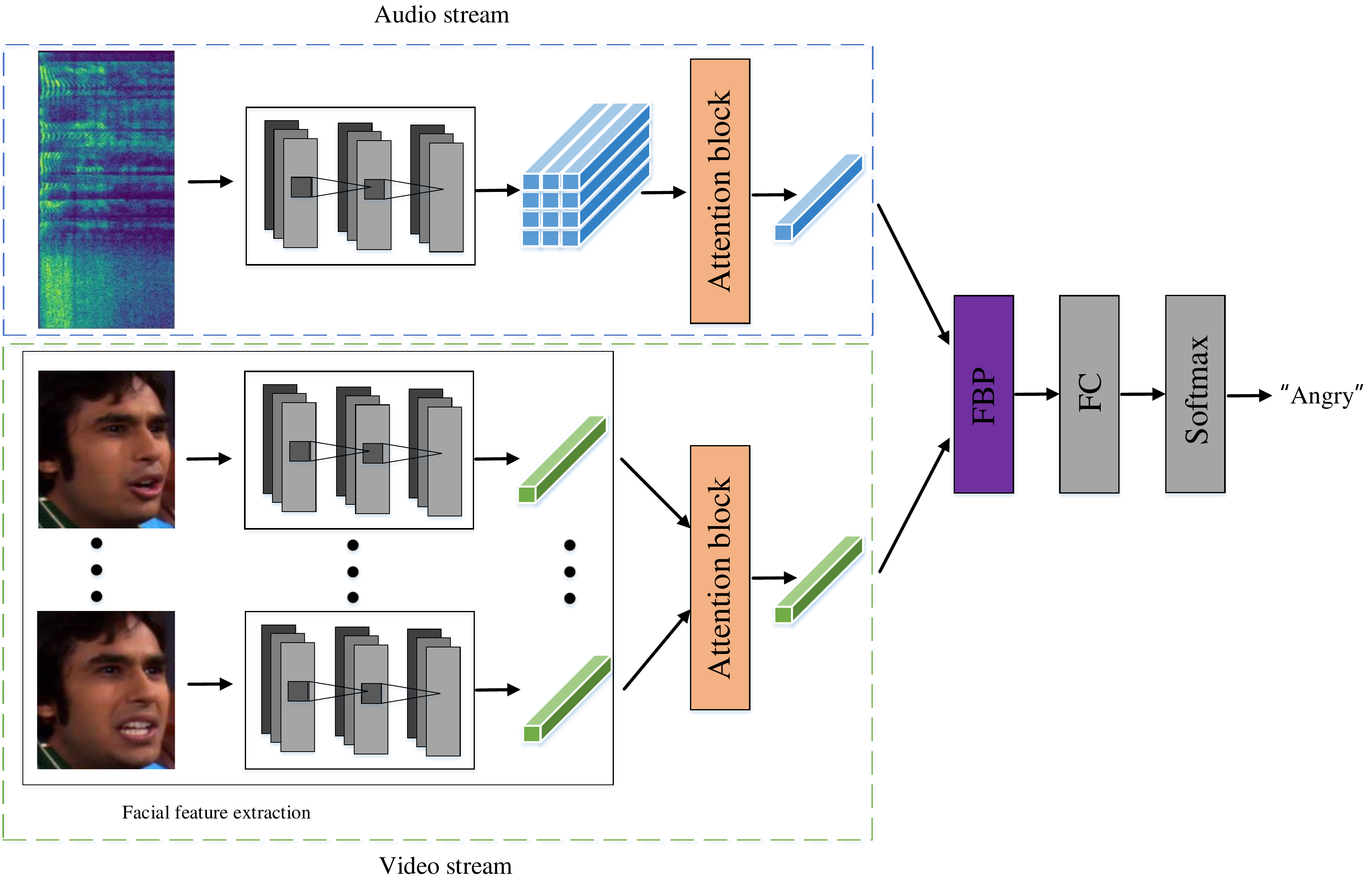}
\caption{The pipeline of our attention guided factorized bilinear pooling system for audio-video emotion recognition.  }
\label{fig_system_architecture}
\end{figure*}

As another carrier of emotion, speech has been widely investigated for emotion recognition for decades. Compared with FER, the traditional audio-based approaches have similar steps: First, with a left-to-right sliding window, sequential acoustic features are extracted from raw speech waveform. Then, based on various statistical functions, for each sentence,  a global feature vector is obtained by combining all corresponding frame-level features. For example, the 1582 dimensions feature extracted using the OpenSmile toolkit~\cite{eyben2010opensmile}, which is mostly used by participators of EmotiW~\cite{dhall2018emotiw}. Finally, the utterance vector is fed to a classifier. With the emerging deep learning, the state-of-the-art classifiers are always CNNs or LSTMs~\cite{satt2017efficient},~\cite{mirsamadi2017automatic},~\cite{tang2018end}. In~\cite{zhang2018attention}, we proposed a novel attention based fully convolutional network for this task.

Like human beings simultaneously depend on auditory and visual systems to apperceive the world, machine can also greatly benefit from auditory and visual information for emotion recognition. The most common strategy is to fuse results from two separate systems, i.e., audio system and visual system, which is referred as decision-level fusion. The decision-level fusion ignores the interaction and correlation between the features from different modalities, which usually has a limited improvement when one modality is much better than the other.
As an improved strategy, in middle layers, using linear fusion (concatenation or element-wise addition) for audio feature and visual feature from two parallel sub-systems can alleviate the shortage in a degree. However, since the feature distributions of different modalities vary dramatically, there is a lack of fully capturing complex association between audio and video for these integrated features obtained by such linear models. In contrast to linear models, bilinear pooling~\cite{tenenbaum2000separating} has recently been used to integrate different CNN features for fine-grained image recognition~\cite{lin2015bilinear}.
Inspired by the multi-modal factorized bilinear pooling (MFB) for visual question answering (VQA) in~\cite{yu2017multi}, we introduce factorized bilinear pooling (FBP), an improved version of bilinear pooling to decrease the computational complexity, to integrate the features of audio and video. Specifically, the features are selected through the embedded attention mechanism from respective modalities. The whole pipeline can be completed in a neural network, which is why the proposed approach is named as deep fusion.

This paper is organized as follows. In section~\ref{section:method}, the proposed architecture is introduced. In section~\ref{section:exp}, the experimental results are reported and analysed. Finally, we make a conclusion in section~\ref{section:conclusion}.

\section{The Proposed Architecture}
\label{section:method}
In\cite{zhang2018attention}, we proposed a novel attention based fully convolutional neural network for audio emotion recognition. The proposed attention mechanism helps the model focus on the emotion-relevant regions in speech spectrogram. It is intuitively obvious that the similar strategy can be adopted in videos to detect the emotion-relevant frames. Furthermore, we first extract emotion-relevant features for audio and video respectively through two parallel streams. And then the separate salient features are fused in a FBP block to deeply capture the association between modalities for final emotion prediction. The overall system architecture is shown in Fig.~\ref{fig_system_architecture}.

\subsection{Audio Stream}

\begin{figure}[!ht]
\centering
\includegraphics[width=1.0\linewidth]{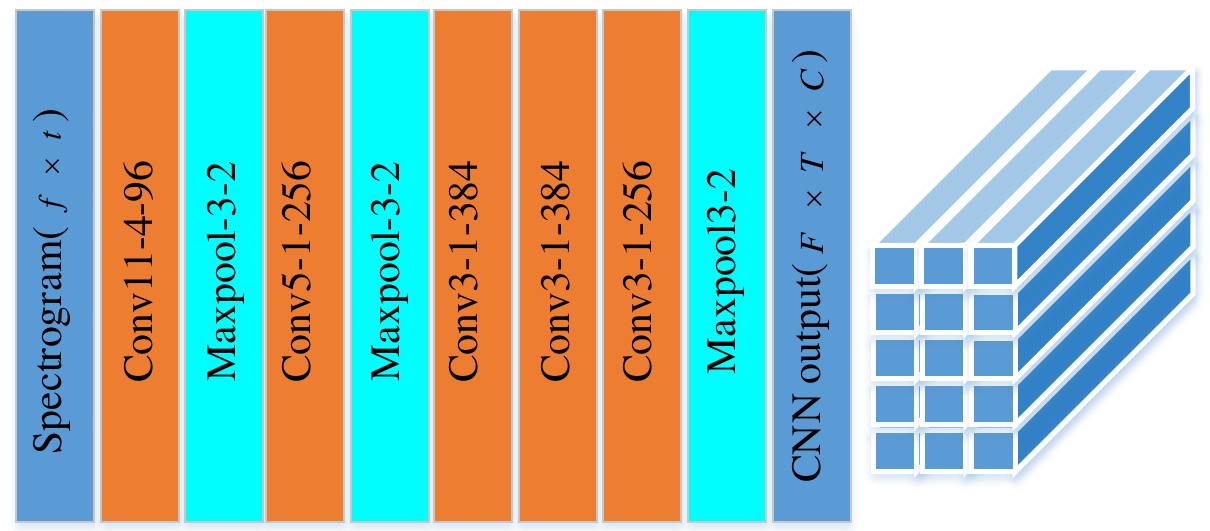}
\caption{The AlexNet based audio encoder. The parameters of the convolutional layer are denoted as  ``Conv(kernel size)-[stride size]-[number of channels]''. The maxpooling layer is denoted as ``Maxpool-[kernel size]-[stride size]''. For brevity, the local response normalization and ReLU activation function are not shown.}
\label{audio_encoder}
\end{figure}

As illustrated in Fig.~\ref{fig_system_architecture}, the audio stream directly handles the speech spectrogram by using stacked convolutional layers followed by an attention block. Without handcrafted feature extraction, CNN has been widely used in speech emotion recognition (SER)~\cite{mirsamadi2017automatic},~\cite{trigeorgis2016adieu},~\cite{aldeneh2017using}. The basic components of CNN  are convolution, pooling and activation layers. The parameters in a convolutional layer include: input channels, output channels, kernel size and stride. Each kernel is a filter with size smaller than the input feature map. The characteristics of kernel make it operate on a local region of input rather than the whole feature map. On a given feature map, the parameters of the kernel are shared to detect the certain pattern in different locations and to reduce the complexity of network. Aiming to remove noise and extract robust features, a pooling layer usually conducts an average or max operation. An activation layer is element-wise nonlinear function usually following a convolutional layer. 

The typical CNNs, including AlexNet ~\cite{krizhevsky2012imagenet}, VGGNet~\cite{simonyan2014very}, and ResNet~\cite{he2016deep} take a fixed-size input due to the limitation of fully connected layers. Considering the loss of information caused by the fixed-size input, we proposed a fully convolutional network to handle variable-length speech in~\cite{zhang2018attention}. In this study, the same is used as audio encoder, which is shown in Fig.~\ref{fig_system_architecture}.  We turn the AlexNet into a fully convolutional network by simply removing its fully connected layers. The details of the audio encoder are shown in  Fig.~\ref{audio_encoder}. All the convolutional layers are followed by a ReLU activation function, and the first two layers are equipped with a local response normalization. 

Assuming the output of the audio encoder is a 3-dimensional array of size $F\times T\times C$, where the $F$ and $T$ correspond to the frequency and time domains of spectrogram and $C$ is the number of channels. We consider the output as a variable-length grid of $L$ elements, $L=F\times T$. Each of the elements is a $C$-dimensional vector corresponding to a region of speech spectrogram, represented as $\boldsymbol{a}_{i}$. Therefore, the whole audio utterance can be represented as a set:
\begin{equation}
\label{FCN_output}
A =\{ \boldsymbol{a}_{1},\cdots,\boldsymbol{a}_{L}\}  ,  \boldsymbol{a}_{i} \in \mathbb{R}^C
\end{equation}

Intuitively, not all time-frequency units in set $A$ contribute equally to the emotion state of the whole utterance. Based on this assumption, we introduce attention mechanism to extract the elements that are important to the emotion of the utterance. And then the sum of weighted elements in the set $A$ represents the audio-based emotional feature. We formulate the attention block as:
\begin{equation}
\label{audio_attention_trans_inner}
e_{i} = \boldsymbol{u}^{T} \tanh(\boldsymbol{W}\boldsymbol{a}_{i}+\boldsymbol{b})
\end{equation}
\begin{equation}
\label{audio_scale_softmax}
\alpha_{i}=\frac{\exp(\lambda e_i)}{\sum_{k=1}^{L}\exp(\lambda e_k)}
\end{equation}
\begin{equation}
\label{audio_weighted_sum}
\boldsymbol{a} = \sum_{i=1}^{L}\alpha_{i}\boldsymbol{a}_{i}
\end{equation}

First, an element $\boldsymbol{a}_{i}$ is fed to a fully connected layer followed by a $\tanh$ to obtain a new representation of $\boldsymbol{a}_{i}$. Then we measure the importance weight $e_{i}$, of the element $\boldsymbol{a}_{i}$ by the inner product between the new representation of $\boldsymbol{a}_{i}$ and a learnable vector $\boldsymbol{u}$. After that, the normalized importance weight $\alpha_{i}$ is calculated through the softmax function. Finally, vector $\boldsymbol{a}$ is computed by the weighted sum of the elements in the set $A$ with importance weights as the audio-based emotional feature. In Eq.(~\ref{audio_scale_softmax}), $\lambda$ is a scale factor which controls the uniformity of the importance weights. $\lambda$ ranges from 0 to 1. If $\lambda=1$, the scaled-softmax becomes the commonly used softmax function. If $\lambda=0$, the vector $\boldsymbol{a}$ will be an average vector of the set $A$, which means all the time-frequency units have the same importance weights to the final utterance audio vector.

\begin{figure}[!ht]
\centering
\includegraphics[width=1.0\linewidth]{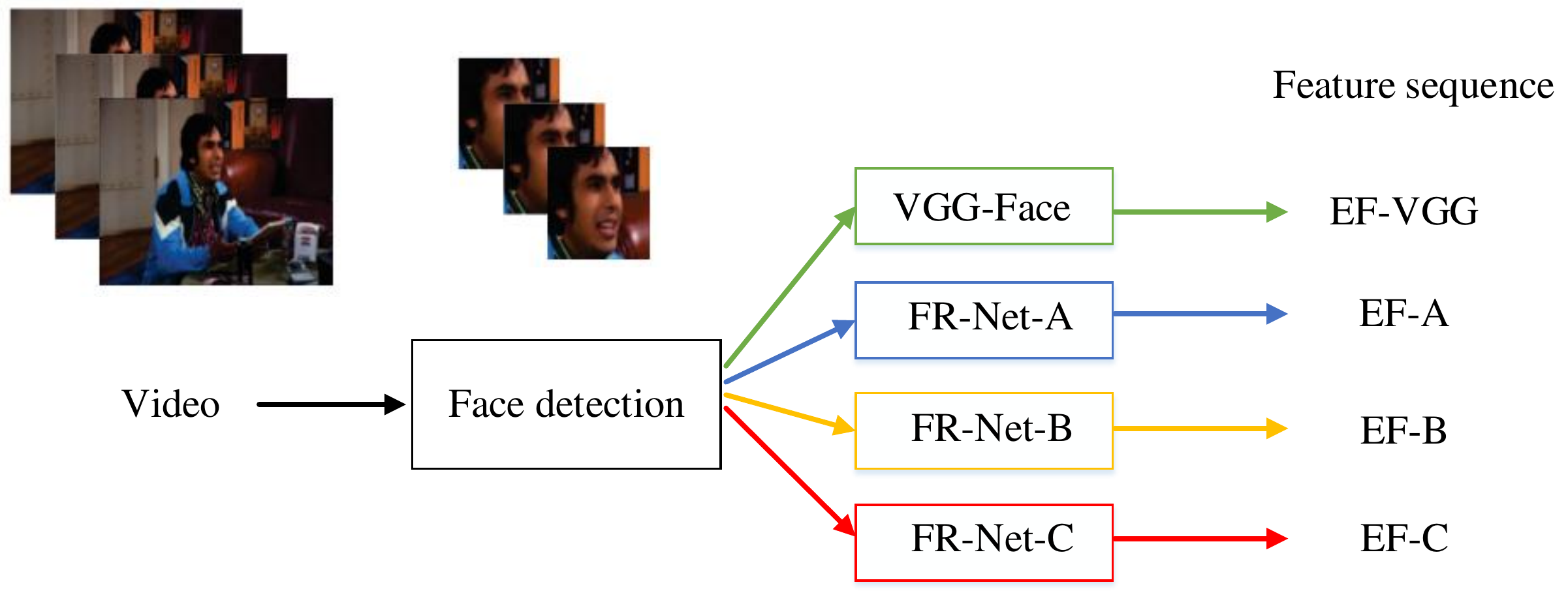}
\caption{The pipeline of the facial feature extraction.}
\label{video_feature_extraction}
\end{figure}

\subsection{Video Stream}
In\cite{knyazev2018leveraging}, four kinds of facial features are proposed for emotion recognition. The pipeline of the feature extraction is illustrated in Fig.~\ref{video_feature_extraction}.
\paragraph{Face detection}
The dlib face detector~\cite{king2009dlib} is used to extract and align face images from FER2013 database\cite{carrier2013fer} and EmotiW video frames. For the frames in which no faces are found, the entire images are used.
\paragraph{FER2013 fine-tuning}
To extract the emotion-related features from a face image, four deep convolutional neural networks, i.e., VGG-Face~\cite{parkhi2015deep}, and three proprietary state-of-the-art face recognition networks which are notated as FR-Net-A, FR-Net-B, FR-Net-C, are used. All models are fine-tuned on FER2013 database to make them emotion-relevant.
\paragraph{Emotional feature extraction}
Features for all frames are computed using all four networks. For VGG-Face, a 4096-dimensional $fc6$ feature is selected. For other networks, the outputs of the last layers are used.
 
 In this paper, we notate them as EF-VGG, EF-A, EF-B, EF-C respectively. The visual feature sequence of a $L$-frames video can be represented as:
 \begin{equation}
\label{face_feature_seq}
V =\{ \boldsymbol{v}_{1},\cdots,\boldsymbol{v}_{L}\}  ,  \boldsymbol{v}_{i} \in \mathbb{R}^C
\end{equation}
where $\boldsymbol{v}_{i}$ represents the facial feature of frame $i$. And $C$ is the feature dimension.

Similarly, not all frames of a video contribute equally to the emotional state. Therefore, we adapt the attention mechanism to compute the importance weights for all frames. And then all the weighted visual features are added to represent the video-based emotional feature. Before entering into the attention block, the dimension reduction is conducted to decrease computational complexity and to relieve over-fitting. The formulae are listed below:
\begin{equation}
\label{video_attention_trans_inner}
\tilde{\boldsymbol{v}}_{i}=\boldsymbol{W}\boldsymbol{v}_{i}+\boldsymbol{b}
\end{equation}
\begin{equation}
\tilde{e}_{i} = \tilde{\boldsymbol{u}}^{T} \tanh(\tilde{\boldsymbol{v}}_{i})
\end{equation}
\begin{equation}
\label{video_scale_softmax}
\tilde{\alpha}_{i}=\frac{\exp(\lambda \tilde{e}_i)}{\sum_{k=1}^{L}\exp(\lambda \tilde{e}_k)}
\end{equation}
\begin{equation}
\label{video_weighted_sum}
\tilde{\boldsymbol{v}} = \sum_{i=1}^{L}\tilde{\alpha}_{i}\tilde{\boldsymbol{v}}_{i}
\end{equation}
where $\tilde{\boldsymbol{v}}_{i}$ is a new low-dimension representation of frame $i$. And $\tilde{\boldsymbol{v}}$ is the video-based emotional feature, which is replaced with $\boldsymbol{v}$ for simplifying in the following subsections. 

\subsection{Audio-video Factorized Bilinear Pooling}

\begin{figure}[!ht]
\centering
\includegraphics[width=0.9\linewidth]{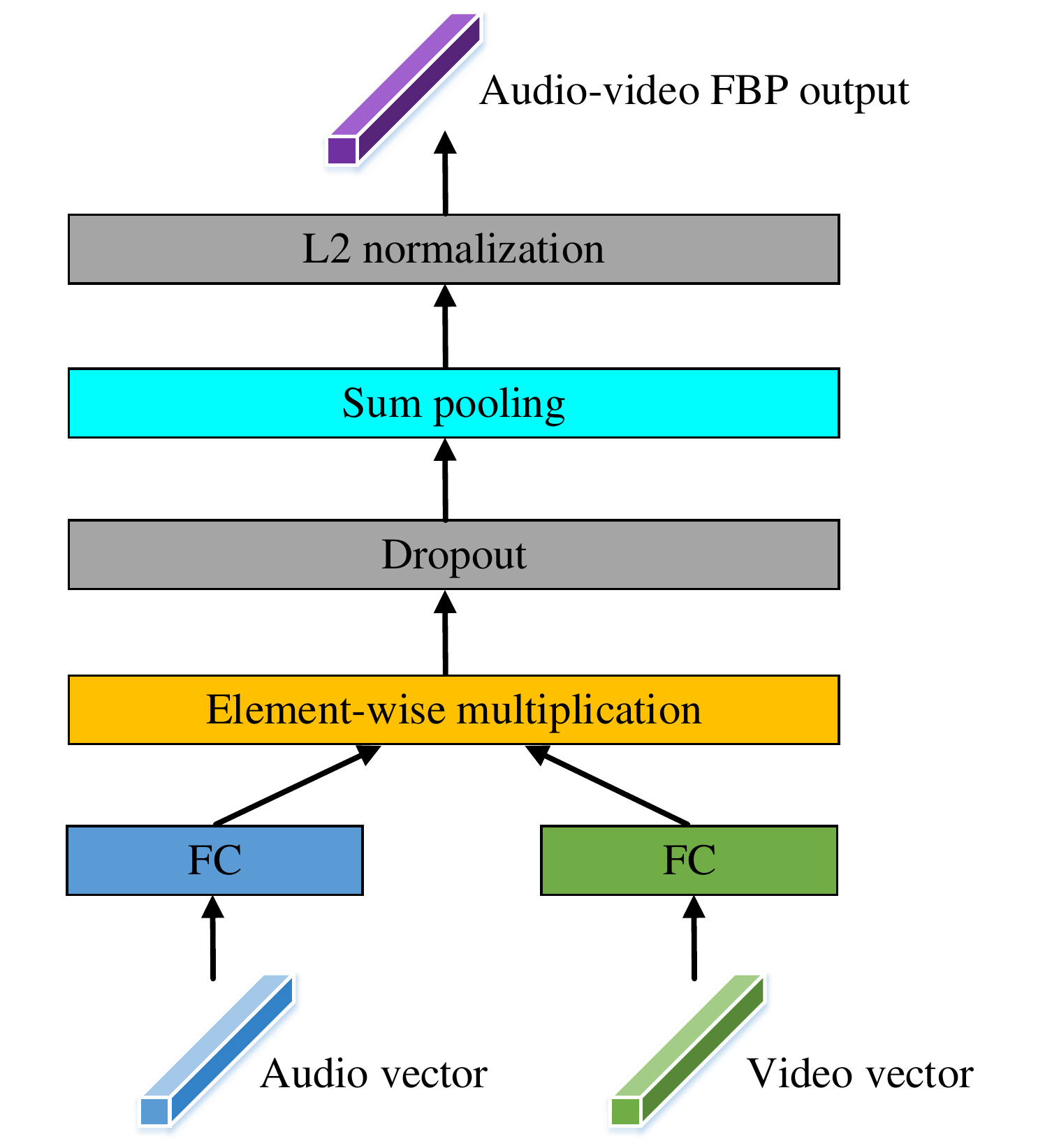}
\caption{The flowchart of the audio-video factorized bilinear pooling operation.}
\label{audio_video_FBP}
\end{figure}

Given two feature vectors in different modalities: the audio feature vector $\boldsymbol{a}\in\mathbb{R}^m$ and video feature vector $\boldsymbol{v}\in\mathbb{R}^n$, the simplest multi-modal bilinear pooling is defined as follows:
\begin{equation}
\label{bilinear_pooling}
{z}_{i}=\boldsymbol{a}^{T}\boldsymbol{W}_{i}\boldsymbol{v}
\end{equation} 
where $\boldsymbol{W}_{i}\in\mathbb{R}^{m\times n}$ is a projection matrix, ${z}_{i}\in\mathbb{R}$ is the output of bilinear pooling. To obtain an $o$-dimensional output $\boldsymbol{z}=[z_{1},\cdots,z_{o}]$, a $\boldsymbol{W}=[\boldsymbol{W}_{1},\cdots,\boldsymbol{W}_{o}]\in\mathbb{R}^{m\times n\times o}$ is need to be learned. Although bilinear pooling can effectively capture the pairwise interactions between the multi-modal features, it also introduces huge number of parameters which lead to a high computational cost and a risk of over-fitting\cite{yu2017multi}.

 Inspired by the matrix factorization tricks\cite{li2017factorized},\cite{rendle2010factorization}, the projection matrix $\boldsymbol{W}_{i}$ in Eq.(\ref{bilinear_pooling}) can be factorized into two low-rank matrices:
 \begin{equation}
 \label{factorization}
 \begin{aligned}
 {z}_{i}&=\boldsymbol{a}^{T}\boldsymbol{U}_{i}\boldsymbol{V}_{i}^{T}\boldsymbol{v}\\&=\sum_{d=1}^{k}\boldsymbol{a}^{T}\boldsymbol{u}_{d}\boldsymbol{v}_{d}^{T}\boldsymbol{v}\\&=\mathbbm{1}^{T}(\boldsymbol{U}_{i}^{T}\boldsymbol{a}\circ \boldsymbol{V}_{i}^{T}\boldsymbol{v})
 \end{aligned}
 \end{equation}
 where $k$ is the latent dimension of the factorizing matrices $\boldsymbol{U}_{i}=[\boldsymbol{u}_{1},\cdots,\boldsymbol{u}_{k}]\in\mathbb{R}^{m\times k}$ and $\boldsymbol{V}_{i}=[\boldsymbol{v}_{1},\cdots,\boldsymbol{v}_{k}]\in\mathbb{R}^{n\times k}$, $\circ$ represents the element-wise multiplication of two vectors, and $\mathbbm{1}\in\mathbb{R}^{k}$ is an all-1 vector.
 
 To obtain the output feature vector $\boldsymbol{z}$ by Eq.(\ref{factorization}), two 3-D tensors $\boldsymbol{U}=[\boldsymbol{U}_{1},\cdots,\boldsymbol{U}_{o}]\in\mathbb{R}^{n\times k \times o}$ and $\boldsymbol{V}=[\boldsymbol{V}_{1},\cdots,\boldsymbol{V}_{o}]\in\mathbb{R}^{n\times k \times o}$ are need to be learned. The $\boldsymbol{U}$ and $\boldsymbol{V}$ can be reformulated as 2-D matrices $\tilde{\boldsymbol{U}}\in\mathbb{R}^{m\times ko}$ and $\tilde{\boldsymbol{V}}\in\mathbb{R}^{n\times ko}$ respectively by using reshape operation. Accordingly, we have:
 \begin{equation}
 \label{sumpooling}
 \boldsymbol{z}={\rm SumPooling} (\tilde{\boldsymbol{U}}^{T}\boldsymbol{a}\circ\tilde{\boldsymbol{V}}^{T}\boldsymbol{v},k)
 \end{equation} 
 where $\tilde{\boldsymbol{U}}^{T}\boldsymbol{a}$ and $\tilde{\boldsymbol{V}}^{T}\boldsymbol{v}$ are implemented by feeding $\boldsymbol{a}$ and $\boldsymbol{v}$ to fully connected layers without biases respectively, and the function ${\rm SumPooling} (\boldsymbol{x},k)$ apply sum pooling within a series of non-overlapped windows to $\boldsymbol{x}$. We indicate the Eq.(\ref{sumpooling}) as the factorized bilinear pooling (FBP). 
 
The detailed procedures of audio-video FBP block are illustrated in Fig.~\ref{audio_video_FBP}. Dropout is adopted to prevent over-fitting. The $l2$-normalization ($\boldsymbol{z}\leftarrow \boldsymbol{z}/\|\boldsymbol{z}\|$) is used after FBP to normalize the energy of $\boldsymbol{z}$ to 1, since the magnitude of the output varies dramatically due to the introduced element-wise multiplication.

\section{Experiments}
\label{section:exp}
\subsection{Database}
\label{section:data_feature}
We validate our method in the AFEW8.0 database, which is used in the audio-video sub-challenge of the EmotiW2018. The AFEW database is collected from films and TV series to simulate the real world. It has seven categories: Angry, Disgust, Fear, Happy, Neutral, Sad, Surprise. There are 773 videos and corresponding audios in the training set, 383 in the validation set, and 653 in the test set.

\subsection{Video System}

\begin{figure*}[htbp]
\centering
\includegraphics[width=1.0\linewidth]{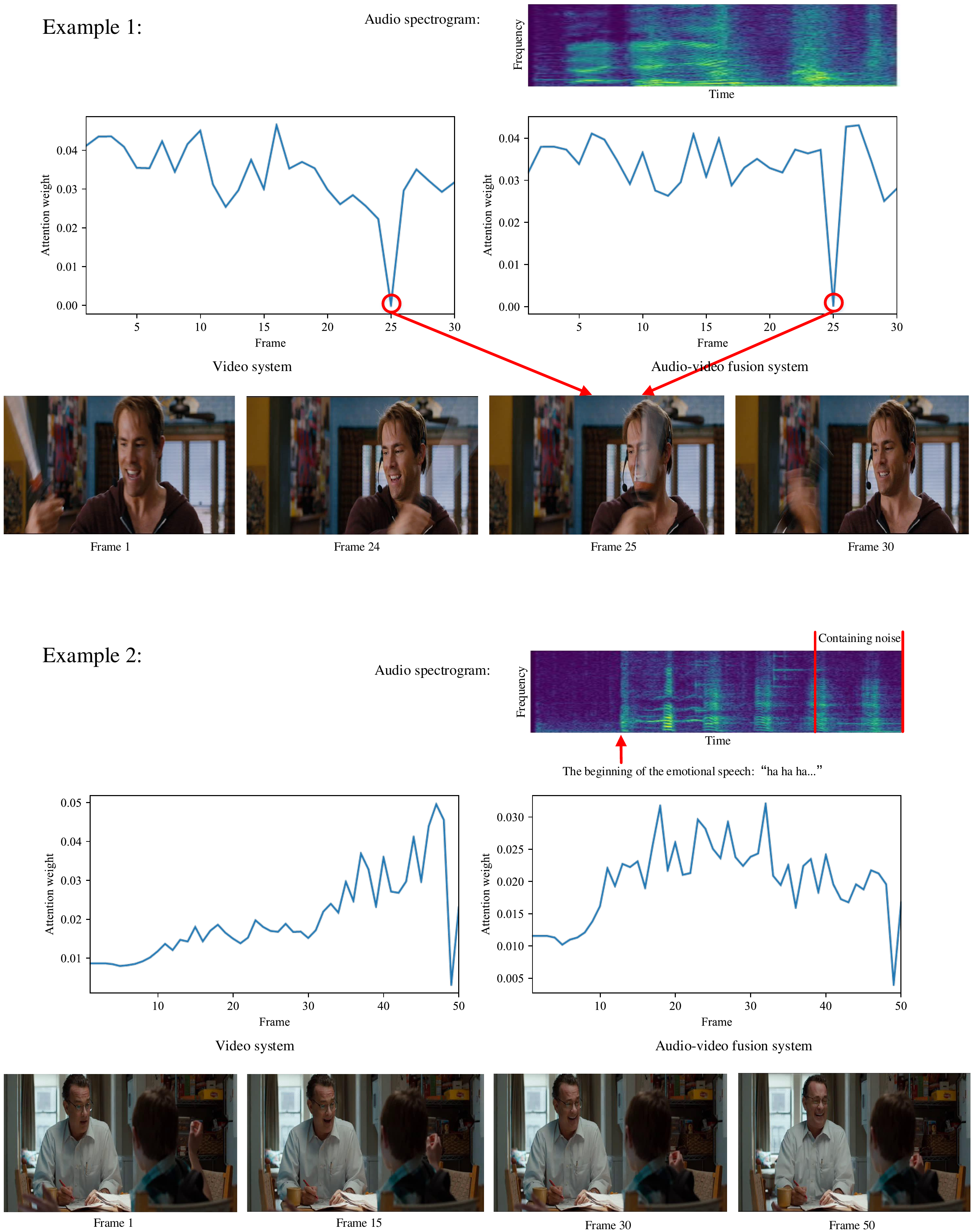}
\caption{The attention weight curves of two randomly selected video examples. For each example, the left curve shows the changing of attention weight along frames in the video system while the right curve derives from the audio-video fusion system. The corresponding audio spectrograms are plotted above the right curves for audio-video analysis. The images are partial frames from the video. }
\label{attention_weight_example}
\end{figure*}

\begin{table}[htbp]
\caption{the classifying accuracy in the validation set of different visual features.}
\label{tab:video_only_acc}
\begin{center}
\begin{tabular}{|c|c|c|}
\hline
Feature &\multicolumn{2}{|c|}{ $\lambda$}\\
\cline{2-3}
name & $0$ & $1$\\
\hline
EF-VGG &$46.74\%$& $47.78\%$ \\
\hline
EF-A &$48.82\%$& $49.87\%$\\
\hline
EF-B &$\boldsymbol{51.44\%}$&$\boldsymbol{51.96\%}$\\
\hline
EF-C &$45.69\%$& $44.91\%$\\
\hline
\end{tabular}
\end{center}
\end{table}

The effectiveness of audio stream for emotion recognition has been proven in~\cite{zhang2018attention}. In order to verify that the attention mechanism is also effective for visual sequences, we first design a system in which only video is used, i.e., we simply remove the audio stream and the FBP block in Fig.~\ref{fig_system_architecture}, and the video vector $\boldsymbol{v}$ is fed to a fully connected layer for classifying. The  scale factor $\lambda$ in the attention block controls the uniformity of the importance weights of the frames in a video. If $\lambda=0$, all the frames contribute the same importance, in which case the attention block equals to a mean pooling for all the frames.

We exam the effectiveness when EF-VGG, EF-A, EF-B, EF-C features are used as input to system respectively. The classification accuracies of the validation set for each case are  listed in Table~\ref{tab:video_only_acc}. From the Table~\ref{tab:video_only_acc}, it can be observed that the systems using EF-B as the feature achieve higher accuracies whether $\lambda$ equals $0$ or $1$, which means the EF-B is more emotion-relevant than other three features. Therefore, the EF-B is selected as the feature for video stream in the audio-video fusion system. Comparing the accuracies between $\lambda=0$ and $\lambda=1$, we conclude that the systems with attention mechanism usually outperform the systems with mean pooling except for the EF-C. The systems using EF-C usually get the lowest accuracies, which means the EF-C feature contains more ``noise'' for this task than others. It is more difficult to detect the emotion-relevant frames  for the attention block of the EF-C system. 

The EF-B system with attention mechanism achieves an accuracy of $54.36\%$ in the test set, which outperforms 15 teams of the total 32 teams in the EmotiW2018. Please note that the systems of participators usually combine multiple audio models and video models, and more training data are used, while the EF-B system is a single model trained with video data only.

To better demonstrate the effect of attention mechanism, we plot the attention weight of each frame for two randomly picked examples from the validation set in Fig.~\ref{attention_weight_example}. For each example, the left curve is the attention weight of each frame in video system. The higher value of the attention weight, the more important the frame is. The example 1 is a video about a man waving a stick. At frame 25, the face of the man is blocked by the waving stick, which means that the frame 25 is unreliable for classifying. And the attention weight of the frame 25 is significantly lower than other frames, which is reasonable. There is a man laughing more and more happy to a boy in example 2. So the increasing trend of the curve is consistent  with the visual information.

\begin{table}[htbp]
\caption{Overall comparison of  different systems.}
\label{tab:concat_vs_FBP}
\begin{center}
\begin{tabular}{|c|c|c|}
\hline
System&Single model&ACC. in the test set\\
\hline
Video-only&Yes&$54.36\%$\\
\hline
Simple concatenation&Yes&$55.28\%$\\
\hline
FBP&Yes&$58.04\%$\\
\hline
FBP (+validation set)&Yes&$60.64\%$\\
\hline
4 FBPs (+validation set)&No&$\boldsymbol{62.48\%}$\\
\hline
CNN+LMED+LSTM~\cite{liu2018multi}&No&$61.87\%$\\
\hline
\end{tabular}
\end{center}
\end{table}

\begin{figure}[!ht]
\centering
\includegraphics[width=1.0\linewidth]{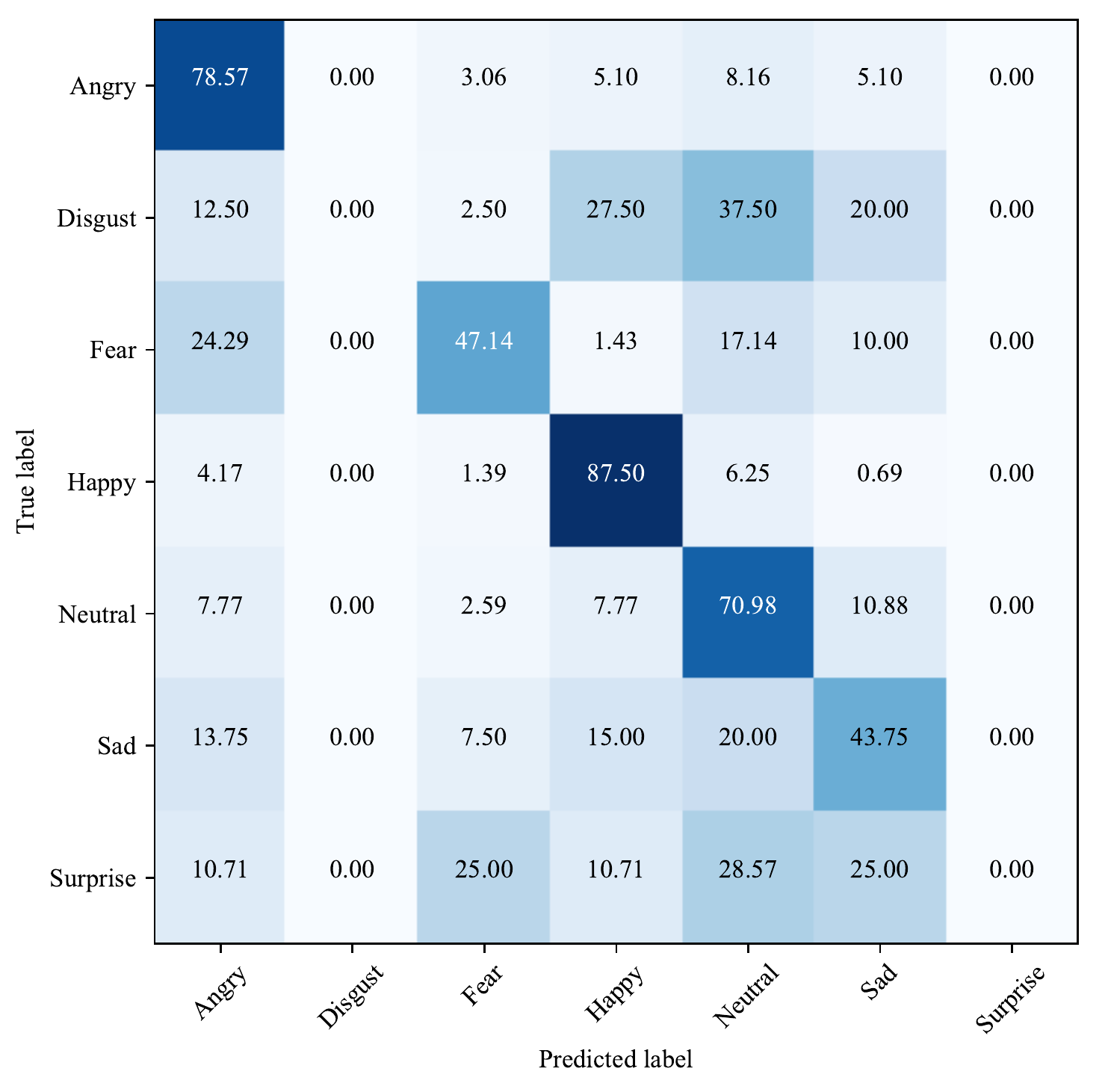}
\caption{The confusion matrix of the audio-video FBP system in the test set.}
\label{confusion_matrix}
\end{figure}

\subsection{Audio-video Fusion System}

For the audio stream, the process in~\cite{zhang2018attention} is applied to extract the audio feature from raw waveform. First, a sequence of overlapping Hamming windows are applied to the speech waveform, with window shift set to 10 msec, and window size set to 40 msec. Then, for each audio frame, we calculate a discrete fourier transform (DFT). Finally the 200-dimensional low-frequency part of the spectrogram is used as the input to the audio stream.

For the video stream, the EF-B is used as the visual feature. A standard normalization is conducted before the visual feature is fed to the network.

In the audio-video fusion experiment, we set the $\lambda=0$ in the attention block of the audio stream, $\lambda=1$ for the video stream. For the FBP block, the $o$ is $128$, the $k$ is $4$, and the dropout probability equals to $0.3$. 

In Table~\ref{tab:concat_vs_FBP}, we compare the accuracies of different fusion strategies in the test set. The concatenation represents simple concatenation of audio and video vectors, replacing the FBP block in Fig.~\ref{fig_system_architecture}. From Table~\ref{tab:concat_vs_FBP}, the audio signal can help video improve performance (from $54.36\%$ to $55.28\%$). And compared with simply concatenating the feature vectors, the FBP obtains a $2.76\%$ improvement, indicating that the FBP can fuse the audio and video information more deeply. 

To analyze how the audio influences the video, we plot the attention weights for all frames of video in the audio-video FBP system in Fig.~\ref{attention_weight_example} (the right curve for each example). The corresponding audio spectrograms are plotted above the curves for the audio-video analysis. For example 1, compared with the left curve, the right curve has less fluctuation in other frames excluding the frame 25, which means the system is more stable and more confident for the unreliability of the frame 25. For example 2, the increasing trend of attention weight disappears. Instead, the weights of middle frames are larger than the beginning and ending frames. Based on the observation of the corresponding audio spectrogram, it is obvious that the  frames with high attention weights are consistent to the emotional speech: ``ha ha ha...'', and the attention weights decrease at the ending frames due to the background noise of the corresponding segment in audio. It shows that the attention block in the video stream is influenced by the audio signal due to the FBP fusion and joint training, i.e., the audio and video are fused deeply. 

Finally, we add the validation set for training as the previous teams in EmotiW~\cite{hu2017learning},~\cite{fan2016video}. The accuracy in the test set is further improved to $60.64\%$. In Table~\ref{tab:concat_vs_FBP}, we also list the previous state-of-the-art accuracy ($61.87\%$) in AFEW database~\cite{liu2018multi}, which is achieved by combining five visual models and two audio models. We achieve a comparable result with a single model, which proves the effectiveness of the proposed system. Moreover, we mean four independent randomly initialized audio-video FBP systems (indicated as 4 FBPs), and the validation set is also added for training, to achieve an accuracy of $62.48\%$ in the test set, which outperform the state-of-the-art result.

For the further analysis, we illustrate the confusion matrix of the audio-video FBP system in the test set in Fig.~\ref{confusion_matrix}. From the confusion matrix, an observation can be found: for emotions that have obvious characteristics are classified more correctly such as angry, happy and neutral. Meanwhile, the emotions such as disgust and surprise are difficult for our model to classify on account of the weak expression and natural confusion with other emotions.

\section{Conclusion}
\label{section:conclusion}  
In this paper, we introduce factorized bilinear pooling (FBP) to deeply integrate the features of audio and video for audio-video emotion recognition. Specifically, the features are selected through the embedded attention mechanism from respective modalities to obtain the emotion-related regions. Furthermore, visualization analysis of the attention weights helps to understand the effectiveness of attention mechanism and multi-modal fusion. Compared with the state-of-the-art approach, the proposed approach can achieve a comparable result with a single model, and make a new milestone with multi-models. For the future work, we plan to further reveal and utilize the relevance of audio and video.  
\section*{Acknowledgment}

The authors would like to thank the organizers of EmotiW for evaluating the accuracies of the proposed systems in the test set of AFEW database.

This work was supported in part by the National Natural Science Foundation of China under Grants U1613211 and 61671422, in part by the Key Science and Technology Project of Anhui Province under Grant 17030901005.
\bibliographystyle{IEEEtran}

\bibliography{mybib}

\begin{thebibliography}{10}
\providecommand{\url}[1]{#1}
\csname url@samestyle\endcsname
\providecommand{\newblock}{\relax}
\providecommand{\bibinfo}[2]{#2}
\providecommand{\BIBentrySTDinterwordspacing}{\spaceskip=0pt\relax}
\providecommand{\BIBentryALTinterwordstretchfactor}{4}
\providecommand{\BIBentryALTinterwordspacing}{\spaceskip=\fontdimen2\font plus
\BIBentryALTinterwordstretchfactor\fontdimen3\font minus
  \fontdimen4\font\relax}
\providecommand{\BIBforeignlanguage}[2]{{%
\expandafter\ifx\csname l@#1\endcsname\relax
\typeout{** WARNING: IEEEtran.bst: No hyphenation pattern has been}%
\typeout{** loaded for the language `#1'. Using the pattern for}%
\typeout{** the default language instead.}%
\else
\language=\csname l@#1\endcsname
\fi
#2}}
\providecommand{\BIBdecl}{\relax}
\BIBdecl

\bibitem{cowie2001emotion}
R.~Cowie, E.~Douglas-Cowie, N.~Tsapatsoulis, G.~Votsis, S.~Kollias, W.~Fellenz,
  and J.~G. Taylor, ``Emotion recognition in human-computer interaction,''
  \emph{IEEE Signal processing magazine}, vol.~18, no.~1, pp. 32--80, 2001.

\bibitem{darwin1998expression}
C.~Darwin and P.~Prodger, \emph{The expression of the emotions in man and
  animals}.\hskip 1em plus 0.5em minus 0.4em\relax Oxford University Press,
  USA, 1998.

\bibitem{tian2001recognizing}
Y.-I. Tian, T.~Kanade, and J.~F. Cohn, ``Recognizing action units for facial
  expression analysis,'' \emph{IEEE Transactions on pattern analysis and
  machine intelligence}, vol.~23, no.~2, pp. 97--115, 2001.

\bibitem{facialexp}
``Facial expression --- {Wikipedia}{,} the free encyclopedia,''
  \url{https://en.wikipedia.org/wiki/Facial_expression}, 2019.

\bibitem{dhall2018emotiw}
A.~Dhall, A.~Kaur, R.~Goecke, and T.~Gedeon, ``Emotiw 2018: Audio-video,
  student engagement and group-level affect prediction,'' in \emph{Proceedings
  of the 2018 on International Conference on Multimodal Interaction}.\hskip 1em
  plus 0.5em minus 0.4em\relax ACM, 2018, pp. 653--656.

\bibitem{dhall2012collecting}
A.~Dhall, R.~Goecke, S.~Lucey, T.~Gedeon \emph{et~al.}, ``Collecting large,
  richly annotated facial-expression databases from movies,'' \emph{IEEE
  multimedia}, vol.~19, no.~3, pp. 34--41, 2012.

\bibitem{ko2018brief}
B.~C. Ko, ``A brief review of facial emotion recognition based on visual
  information,'' \emph{sensors}, vol.~18, no.~2, p. 401, 2018.

\bibitem{hu2017learning}
P.~Hu, D.~Cai, S.~Wang, A.~Yao, and Y.~Chen, ``Learning supervised scoring
  ensemble for emotion recognition in the wild,'' in \emph{Proceedings of the
  19th ACM International Conference on Multimodal Interaction}.\hskip 1em plus
  0.5em minus 0.4em\relax ACM, 2017, pp. 553--560.

\bibitem{fan2018video}
Y.~Fan, J.~C. Lam, and V.~O. Li, ``Video-based emotion recognition using
  deeply-supervised neural networks,'' in \emph{Proceedings of the 2018 on
  International Conference on Multimodal Interaction}.\hskip 1em plus 0.5em
  minus 0.4em\relax ACM, 2018, pp. 584--588.

\bibitem{tran2015learning}
D.~Tran, L.~Bourdev, R.~Fergus, L.~Torresani, and M.~Paluri, ``Learning
  spatiotemporal features with 3d convolutional networks,'' in
  \emph{Proceedings of the IEEE international conference on computer vision},
  2015, pp. 4489--4497.

\bibitem{vielzeuf2017temporal}
V.~Vielzeuf, S.~Pateux, and F.~Jurie, ``Temporal multimodal fusion for video
  emotion classification in the wild,'' in \emph{Proceedings of the 19th ACM
  International Conference on Multimodal Interaction}.\hskip 1em plus 0.5em
  minus 0.4em\relax ACM, 2017, pp. 569--576.

\bibitem{kim2017multi}
D.~H. Kim, M.~K. Lee, D.~Y. Choi, and B.~C. Song, ``Multi-modal emotion
  recognition using semi-supervised learning and multiple neural networks in
  the wild,'' in \emph{Proceedings of the 19th ACM International Conference on
  Multimodal Interaction}.\hskip 1em plus 0.5em minus 0.4em\relax ACM, 2017,
  pp. 529--535.

\bibitem{eyben2010opensmile}
F.~Eyben, M.~W{\"o}llmer, and B.~Schuller, ``Opensmile: the munich versatile
  and fast open-source audio feature extractor,'' in \emph{Proceedings of the
  18th ACM international conference on Multimedia}.\hskip 1em plus 0.5em minus
  0.4em\relax ACM, 2010, pp. 1459--1462.

\bibitem{satt2017efficient}
A.~Satt, S.~Rozenberg, and R.~Hoory, ``Efficient emotion recognition from
  speech using deep learning on spectrograms,'' \emph{Proc. Interspeech 2017},
  pp. 1089--1093, 2017.

\bibitem{mirsamadi2017automatic}
S.~Mirsamadi, E.~Barsoum, and C.~Zhang, ``Automatic speech emotion recognition
  using recurrent neural networks with local attention,'' in \emph{Acoustics,
  Speech and Signal Processing (ICASSP), 2017 IEEE International Conference
  on}.\hskip 1em plus 0.5em minus 0.4em\relax IEEE, 2017, pp. 2227--2231.

\bibitem{tang2018end}
D.~Tang, J.~Zeng, and M.~Li, ``An end-to-end deep learning framework with
  speech emotion recognition of atypical individuals,'' \emph{Proc. Interspeech
  2018}, pp. 162--166, 2018.

\bibitem{zhang2018attention}
Y.~Zhang, J.~Du, Z.~Wang, and J.~Zhang, ``Attention based fully convolutional
  network for speech emotion recognition,'' \emph{arXiv preprint
  arXiv:1806.01506}, 2018.

\bibitem{tenenbaum2000separating}
J.~B. Tenenbaum and W.~T. Freeman, ``Separating style and content with bilinear
  models,'' \emph{Neural computation}, vol.~12, no.~6, pp. 1247--1283, 2000.

\bibitem{lin2015bilinear}
T.-Y. Lin, A.~RoyChowdhury, and S.~Maji, ``Bilinear cnn models for fine-grained
  visual recognition,'' in \emph{Proceedings of the IEEE International
  Conference on Computer Vision}, 2015, pp. 1449--1457.

\bibitem{yu2017multi}
Z.~Yu, J.~Yu, J.~Fan, and D.~Tao, ``Multi-modal factorized bilinear pooling
  with co-attention learning for visual question answering,'' in \emph{Proc.
  IEEE Int. Conf. Comp. Vis}, vol.~3, 2017.

\bibitem{trigeorgis2016adieu}
G.~Trigeorgis, F.~Ringeval, R.~Brueckner, E.~Marchi, M.~A. Nicolaou,
  B.~Schuller, and S.~Zafeiriou, ``Adieu features? end-to-end speech emotion
  recognition using a deep convolutional recurrent network,'' in
  \emph{Acoustics, Speech and Signal Processing (ICASSP), 2016 IEEE
  International Conference on}.\hskip 1em plus 0.5em minus 0.4em\relax IEEE,
  2016, pp. 5200--5204.

\bibitem{aldeneh2017using}
Z.~Aldeneh and E.~M. Provost, ``Using regional saliency for speech emotion
  recognition,'' in \emph{Acoustics, Speech and Signal Processing (ICASSP),
  2017 IEEE International Conference on}.\hskip 1em plus 0.5em minus
  0.4em\relax IEEE, 2017, pp. 2741--2745.

\bibitem{krizhevsky2012imagenet}
A.~Krizhevsky, I.~Sutskever, and G.~E. Hinton, ``Imagenet classification with
  deep convolutional neural networks,'' in \emph{Advances in neural information
  processing systems}, 2012, pp. 1097--1105.

\bibitem{simonyan2014very}
K.~Simonyan and A.~Zisserman, ``Very deep convolutional networks for
  large-scale image recognition,'' \emph{arXiv preprint arXiv:1409.1556}, 2014.

\bibitem{he2016deep}
K.~He, X.~Zhang, S.~Ren, and J.~Sun, ``Deep residual learning for image
  recognition,'' in \emph{Proceedings of the IEEE conference on computer vision
  and pattern recognition}, 2016, pp. 770--778.

\bibitem{knyazev2018leveraging}
B.~Knyazev, R.~Shvetsov, N.~Efremova, and A.~Kuharenko, ``Leveraging large face
  recognition data for emotion classification,'' in \emph{Automatic Face \&
  Gesture Recognition (FG 2018), 2018 13th IEEE International Conference
  on}.\hskip 1em plus 0.5em minus 0.4em\relax IEEE, 2018, pp. 692--696.

\bibitem{king2009dlib}
D.~E. King, ``Dlib-ml: A machine learning toolkit,'' \emph{Journal of Machine
  Learning Research}, vol.~10, no. Jul, pp. 1755--1758, 2009.

\bibitem{carrier2013fer}
P.-L. Carrier, A.~Courville, I.~J. Goodfellow, M.~Mirza, and Y.~Bengio,
  ``Fer-2013 face database,'' \emph{Universit de Montral}, 2013.

\bibitem{parkhi2015deep}
O.~M. Parkhi, A.~Vedaldi, A.~Zisserman \emph{et~al.}, ``Deep face
  recognition.'' in \emph{BMVC}, vol.~1, no.~3, 2015, p.~6.

\bibitem{li2017factorized}
Y.~Li, N.~Wang, J.~Liu, and X.~Hou, ``Factorized bilinear models for image
  recognition,'' \emph{arXiv preprint}, 2017.

\bibitem{rendle2010factorization}
S.~Rendle, ``Factorization machines,'' in \emph{Data Mining (ICDM), 2010 IEEE
  10th International Conference on}.\hskip 1em plus 0.5em minus 0.4em\relax
  IEEE, 2010, pp. 995--1000.

\bibitem{liu2018multi}
C.~Liu, T.~Tang, K.~Lv, and M.~Wang, ``Multi-feature based emotion recognition
  for video clips,'' in \emph{Proceedings of the 2018 on International
  Conference on Multimodal Interaction}.\hskip 1em plus 0.5em minus 0.4em\relax
  ACM, 2018, pp. 630--634.

\bibitem{fan2016video}
Y.~Fan, X.~Lu, D.~Li, and Y.~Liu, ``Video-based emotion recognition using
  cnn-rnn and c3d hybrid networks,'' in \emph{Proceedings of the 18th ACM
  International Conference on Multimodal Interaction}.\hskip 1em plus 0.5em
  minus 0.4em\relax ACM, 2016, pp. 445--450.

\end{thebibliography}

\end{document}